\documentclass[letterpaper, 10 pt, conference]{ieeeconf}  

\IEEEoverridecommandlockouts                              

\usepackage{color}
\usepackage{graphicx}
\usepackage{amssymb}
\usepackage{bm}
\usepackage{amsmath}
\usepackage{multirow}
\usepackage{booktabs}
\usepackage{makecell}
\usepackage{hyperref}
\hypersetup{
        colorlinks=true, linkcolor=black
    }

\title{\LARGE \bf
CA-SpaceNet: Counterfactual Analysis for 6D Pose Estimation in Space}

\author{Shunli Wang$^{1,2\dag}$, Shuaibing Wang$^{1,2\dag}$, Bo Jiao$^{1,2}$, Dingkang Yang$^{1,2}$, \\Liuzhen Su$^{1,2}$, Peng Zhai$^{1,2}$, Chixiao Chen$^{1,2*}$ and Lihua Zhang$^{1,3,2,4*}$ 
\thanks{This work is supported by Shanghai Municipal Science and Technology Major Project (2021SHZDZX0103) and NSFC Grant (61974033).}%
\thanks{$^{1}$Academy for Engineering \& Technology, Fudan University. $^{2}$Engineering Research Center of AI and Robotics, Ministry of Education, China. $^{3}$Jilin Provincial Key Laboratory of Intelligence Science \& Engineering, China. $^{4}$Artifical Intelligence and Unmanned Systems Engineering Research Center of Jilin Province, China}%
\thanks{$^{\dag}$ The first two authors contributed equally to this work.}
\thanks{$^{*}$ Corresponding author, Email: \{lihuazhang, cxchen\}@fudan.edu.cn}%
}

\begin{document}

\maketitle
\thispagestyle{empty}
\pagestyle{empty}

\begin{abstract}
Reliable and stable 6D pose estimation of uncooperative space objects plays an essential role in on-orbit servicing and debris removal missions.
Considering that the pose estimator is sensitive to background interference, this paper proposes a counterfactual analysis framework named CA-SpaceNet to complete robust 6D pose estimation of the space-borne targets under complicated background.
Specifically, conventional methods are adopted to extract the features of the whole image in the factual case.
In the counterfactual case, a non-existent image without the target but only the background is imagined.
Side effect caused by background interference is reduced by counterfactual analysis, which leads to unbiased prediction in final results.
In addition, we also carry out low-bit-width quantization for CA-SpaceNet and deploy part of the framework to a Processing-In-Memory (PIM) accelerator on FPGA.
Qualitative and quantitative results demonstrate the effectiveness and efficiency of our proposed method.
To our best knowledge, this paper applies causal inference and network quantization to the 6D pose estimation of space-borne targets for the first time.
The code is available at \href{https://github.com/Shunli-Wang/CA-SpaceNet}{https://github.com/Shunli-Wang/CA-SpaceNet}.

\end{abstract}

\section{Introduction}

\begin{figure}[t]
              \centering
              \includegraphics[width=\linewidth]{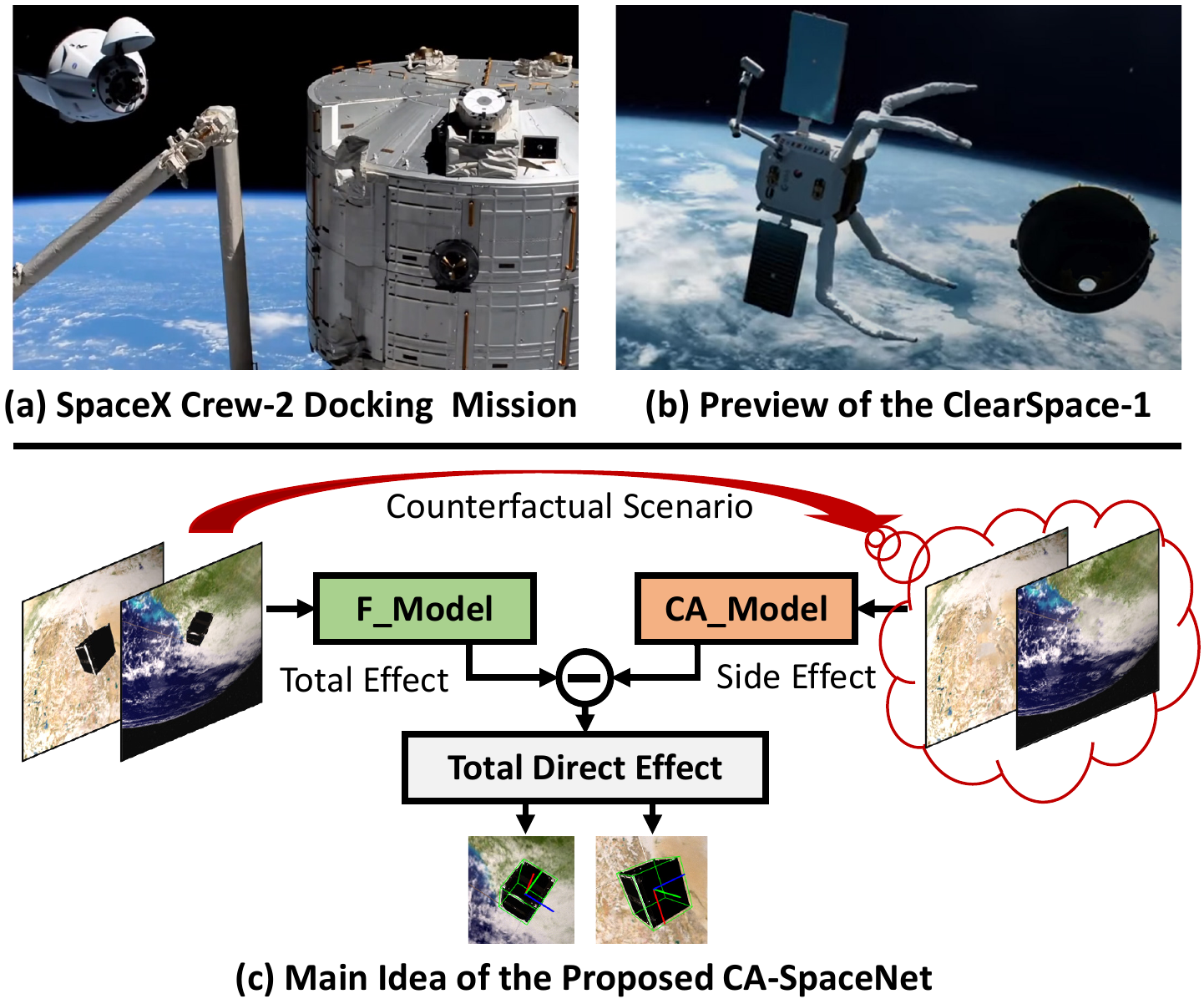}
              \caption{
              Practical applications of the 6D pose estimation in many space missions, such as automatic docking and debris removal.
              (a) shows a screenshot of the docking in NASA’s SpaceX Crew-2 mission performed in 2021.
              (b) demonstrates a preview of the ClearSpace-1 satellite proposed by ESA and ClearSpace company which will be launched in 2025.
              (c) The complicated background of aerial images will interfere with the stability of the 6D pose estimator.
              Therefore, this paper introduces counterfactual analysis to the 6D pose estimation task in space and proposes the CA-SpaceNet framework. 
              By imagining an image without the target (\textit{i.e.}, Side Effect), the CA-SpaceNet can decouple the pure target features (\textit{i.e.}, Total Direct Effect) from the raw features (\textit{i.e.}, Total Effect) through counterfactual analysis to obtain more accurate estimation results.
              }
            \label{fig1}
            \end{figure}

As the eye of the spacecraft, vision-based navigation system is a crucial technology in many unmanned space missions.
6D pose estimation of space-borne objects is the premise of the navigation system.
Fig. \ref{fig1}(a)\&(b) shows two practical applications of 6D pose estimation in space: automatic docking and debris removal missions.
Compared with terrestrial applications, many factors should be considered, such as harsh imaging conditions caused by the lack of atmospheric scattering and limited computing resources and power consumption.
The robust and efficient 6D pose estimator is the key to ensuring the regular operation of on-orbit service. 

In recent years, there have been some studies \cite{iterative_1, PRISMA_1, SPEED, SwissCube, ICCVW_2019} in space engineering and computer vision community to explore 6D pose estimation of space-borne targets.
Although considerable performance has been achieved, many methods directly migrate models from terrestrial to space scene without considering the particularity of space mission. 
In addition, these works mainly focus on the performance improvement of the model and ignore the power consumption and latency of the actual deployment on real spacecraft.

To address these challenges, this paper proposes a Counterfactual Analysis SpaceNet (CA-SpaceNet) framework to handle complicated background information in aerial images.
As demonstrated in Fig. \ref{fig1}(c), the CA-SpaceNet introduces counterfactual analysis to the 6D pose estimation task and constructs factual and counterfactual paths. 
In the factual path, the whole image will be sent to the \textit{F\_Model} to complete feature extraction and results in the factual features (\textit{i.e.}, Total Effect).
In the counterfactual path, an image without target but the background is imagined. This non-existent image will be sent to the \textit{CF\_Model} to complete feature extraction and results in the counterfactual features (\textit{i.e.}, Side Effect).
With the power of causal inference \cite{2016Causal, why}, the CA-SpaceNet can remove the harmful background interference from factual features and generate accurate pose results, which cannot be easily identified by traditional methods.
Secondly, to fill the gap in actual deployment on the low-power consumption hardware of the 6D pose estimator, this paper quantizes the CA-SpaceNet into a low-bit-width model and explores the impact of quantizing different modules on the final performance. 
A part of the quantized network in 3-bit is implemented on FPGA.
Extensive experimental results demonstrate the high performance of the CA-SpaceNet. 
Latency testing on FPGA confirms the efficiency of low-bit-width quantization and the accelerator architecture.

The main contributions of this paper are as follows:
\begin{itemize}
    \item We propose a framework named CA-SpaceNet, which is robust to the interference of complicated background information by introducing counterfactual analysis to the 6D pose estimation task in space.
    \item Our approach outperforms state-of-the-arts on the challenging SwissCube dataset and achieves competitive results on the SPEED dataset.  
    \item We quantize the CA-SpaceNet into a low-bit-width model and deploy a part of the quantized network into a Processing-In-Memory (PIM) chip on FPGA. 
    Low latency proves the feasibility of our method.
\end{itemize}

As far as we know, it is the first time that the causal inference method and network quantization are explored to address the 6D pose estimation task in space.
Robust performance and high efficiency confirm the effectiveness of our method and deployment.

\section{Related Work}

\noindent\textbf{6D Pose Estimation in Space:} 
Monocular-based 6D pose estimation is a fundamental task in computer vision.
According to stages, these methods can be roughly divided into two categories: two-stage and one-stage methods.
Two-stage methods \cite{2017SSD, 2017BB8, 2018Real, PoseCNN} complete the keypoints detection firstly (usually adopt corners of the 3D object bounding box) and then solve the 6D pose by the 3D-to-2D correspondences through a PnP solver \cite{PnP}.
There is no keypoint detection process in one-stage methods \cite{PoseCNN2018, ICRA_Dataset, se3TrackNet}. The pose information of the object will be transformed into unit quaternion and 3-D translation vector, and the model directly regresses these parameters.

6D pose estimation of space-borne targets plays an important role in satellite on-orbit services and on-board vision-based navigation systems \cite{nav_sys_1, nav_sys_2}.
Compared with terrestrial applications, strict navigation and restricted computation resources put forward higher requirements for the pose estimation model in space.
The space engineering community has explored this problem.
Traditional methods \cite{iterative_1, iterative_2, iterative_3} first find an initial state, \textit{i.e.}, \textit{a priori}, and then use the iterative algorithms to solve the best pose solution that minimizes a specific error criterion pose via hand-crafted feature points.
D'Amico \textit{et al.} \cite{PRISMA_1} and Sharma \textit{et al.} \cite{PRISMA_2} proposed some special methods of hand-crafted features to avoid the provision of the initial state based on authentic images captured during the PRISMA mission \cite{PRISMA_1, PRISMA_3}.
Although a series of improvements increase the performance, there is still a huge gap between these optimization-based methods and ideal models.

With the proposal of some large-scale datasets in 6D pose estimation in space \cite{SPEED, ICRA_Dataset, SwissCube}, some deep neural networks (DNNs) based methods \cite{Sharma_2018, ICCVW_2019, ICRA_Dataset, SwissCube} are proposed.
Most of these methods directly modify DNNs to the space scene and do not consider the intrinsic characteristics of space tasks.
However, Hu \textit{et al.} \cite{SwissCube} considered the extensive depth range and proposed the WDR model, which achieved superior performance on the proposed SwissCube dataset.
Inspired by \cite{SwissCube}, the proposed CA-SpaceNet aims to reduce the interference of complex backgrounds and obtain unbiased pose estimation results through a counterfactual analysis strategy.


\noindent\textbf{Counterfactual Analysis:}
Counterfactual analysis originates from psychology, which explores that human beings have the ability to evaluate outcomes that did not occur but could have occurred under different conditions \cite{why}. 
As a powerful way for testing cause-and-effect relationships, counterfactual analysis has been widely used in politics, economics, and epidemiology \cite{Activism, political,medical, Econometrica}.
Recently, the computer vision community has paid more attention to the application of counterfactual analysis in many visual tasks such as long-tailed visual recognition \cite{2020Long}, action anticipation \cite{CA_EAA}, scene graph generation (SGG) \cite{UnbiasedSGG}, and visual question answering (VQA) \cite{VQA}.
Zhang \textit{et al.} \cite{CA_EAA} presented a counterfactual analysis framework for the egocentric action anticipation (EAA) task. Through the construction of factual and counterfactual cases, side effect caused by semantic labels is reduced, which leads to accurate action anticipation results.
The utilization of counterfactual analysis in these methods can improve the performance and interpretability of the model simultaneously.
This paper alleviates the problem that the 6D pose estimation model is easily affected by the background through counterfactual analysis and improves the stability of the model.

\noindent\textbf{Low-bit-width Quantization for DNNs:}
Although DNNs have achieved excellent results in various tasks, the vast computational cost hinders the deployment of these models.
Researchers have to trade off the performance and the cost of deployment, especially in special scenes where the computing resources are strictly limited.
Much work has explored the lightweight of DNNs, such as network pruning \cite{pruning_1, pruning_2}, knowledge distillation \cite{distillation1, distillation2}, and quantization \cite{LSQ,quan_1}.
DNNs run with low precision operations during inference provide power and memory advantages over full precision, and it also benefits low-bit-width artificial intelligence chip design \cite{Jiao, chip_review}.
The main idea of quantization is to map full precision floating-point numbers to lower precision (8-bit or lower) through a quantizer to significantly reduce the amount of floating-point operations (FLOPs) in matrix multiplication.
Most of the existing methods only explore quantization algorithms in the classification task.
In this paper, the proposed CA-SpaceNet is quantized by LSQ-Net \cite{LSQ} and deployed in a low-bit-width PIM accelerator.

\section{Method}





                
            \begin{figure*} 
              \centering
              \includegraphics[width=\linewidth]{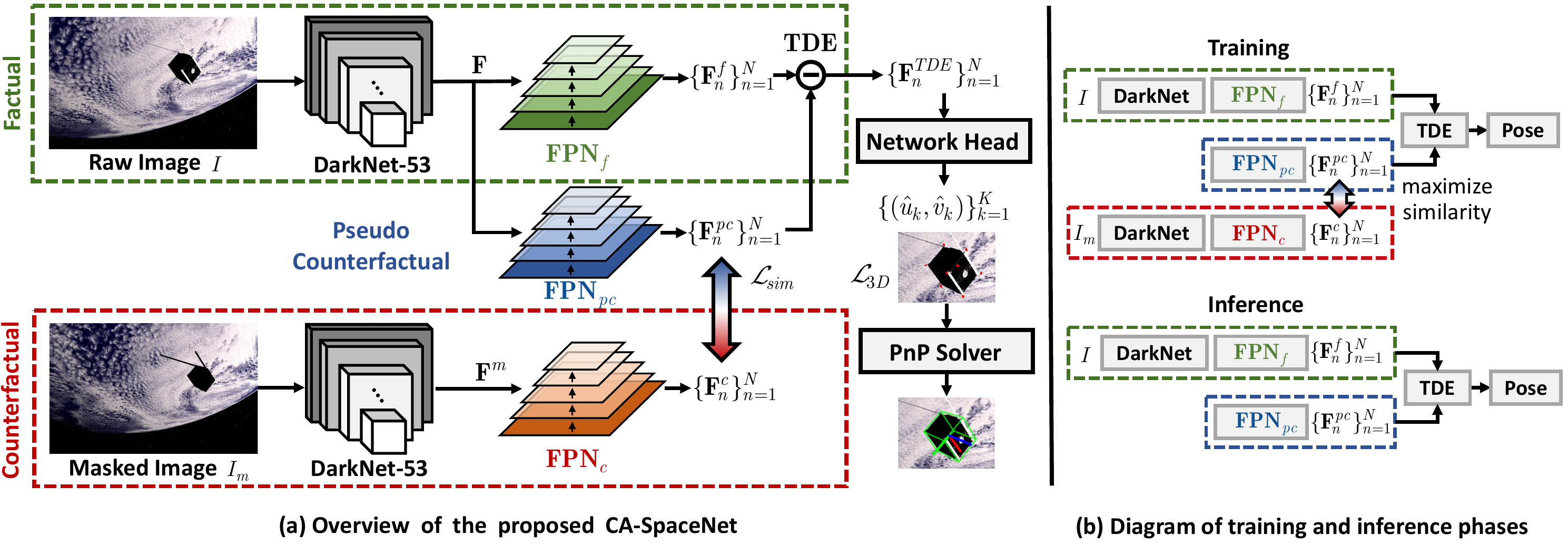}
              \caption{(a) The CA-SpaceNet consists of five stages: 
              2) Three independent feature pyramid networks 
              \textcolor[RGB]{83,129,53}{$ \textbf{FPN}_f $}, 
              \textcolor[RGB]{16,84,151}{$ \textbf{FPN}_{pc} $} 
              and 
              \textcolor[RGB]{192,0,0}{$ \textbf{FPN}_c $}
              with the same structure complete the feature aggregation. 
              For clarity, different colors are assigned to the three paths.
              3) Unbiased feature $\{ \textbf{F}^{TDE}_n \}^{N}_{n=1} $ is obtained by counterfactual analysis.
              4) The keypoint detector regresses the 2D projections of the 3D corners of the satellite's cubic body. 
              5) Finally, the PnP solver is utilized to calculate the 6D pose 
              of the target satellite through 2D-3D correspondences.
              (b) To clearly explain the differences between the training and inference phases, we ignore unnecessary feature arrows. 
              In the training phase, \textcolor[RGB]{16,84,151}{$ \textbf{FPN}_{pc} $} will imitate \textcolor[RGB]{192,0,0}{$ \textbf{FPN}_c $} by maximizing the similarity between $ \{ \textbf{F}^{pc}_{n} \}^{N}_{n=1} $ and $ \{ \textbf{F}^{c}_{n} \}^{N}_{n=1} $, while the whole counterfactual path will be removed during inference. 
              }
            \label{fig2}
            \end{figure*}

\subsection{Overview} 
The network architecture is given in Fig. \ref{fig2}.
The network's input is two images: the first one is the raw image $ I $ with satellite, and the second is the image $ I_m $ with only the background after removing the satellite.
Two DarkNet-53 networks with the same weights are adopted to perform features extraction of $ I $ and $ I_m $, respectively, resulting in $ \textbf{F} \in \mathbb{R}^{C \times H \times W} $ and $ \textbf{F}^m \in \mathbb{R}^{C \times H \times W} $.
After feature extraction, three feature pyramid networks  \textcolor[RGB]{83,129,53}{$ \textbf{FPN}_f $}, \textcolor[RGB]{16,84,151}{$ \textbf{FPN}_{pc} $}, and \textcolor[RGB]{192,0,0}{$ \textbf{FPN}_c $} are constructed to perform counterfactual analysis.
These feature aggregation modules with the same network structure but different weights are the core components of 
\textcolor[RGB]{83,129,53}{\textbf{factual path}}, 
\textcolor[RGB]{16,84,151}{\textbf{pseudo counterfactual path}}, and 
\textcolor[RGB]{192,0,0}{\textbf{counterfactual path}}.
Through counterfactual analysis, the side effect can be decoupled and removed from the total effect to obtain the final TDE, \textit{i.e.}, unbiased feature $\{ \textbf{F}^{TDE}_n \}^{N}_{n=1} $  after weakening background interference, where $N$ denotes the number of layers in FPN.
Finally, the unbiased feature will be sent to an anchor-based keypoint detector. A PnP solver is adopted to predict the final 6D pose of the target satellite.

The rest of this section is organized as follows: 
In subsections \ref{B}, \ref{C}, and \ref{D}, the construction of the factual path, counterfactual path, and pseudo counterfactual path are described in detail, respectively. 
The network quantization method adopted in this paper is briefly reviewed in subsection \ref{E}.
The training and inference processes of the CA-SpaceNet are introduced in subsection \ref{F}.

\subsection{Factual Path} \label{B} 
It can be seen from the ranking of SPEED competition that the methods based on PnP solver are much more stable than the methods of directly estimating 6D pose.
Therefore, this paper adopts the strategy based on a PnP solver. 
In this strategy, the 6D pose estimation task is divided into two subtasks: 2D keypoint detection and PnP problem.
Detailed compositions of the proposed framework are shown in Fig. \ref{fig2}.
The factual path is the central part of the CA-SpaceNet, which refers to the structure of \cite{SwissCube}.
Hu \textit{et al.} \cite{SwissCube} explored the problem of huge changes in the depth range of space-borne objects. However, they directly adopted neural networks to extract features of the whole image without considering the impact of complex background information on 6D pose estimation tasks.
In some general computer vision tasks, \textit{e.g.}, object detection and semantic segmentation, background interference will not cause a significant decrease in performance. 
While in some tasks requiring high precision, such as 6D pose estimation, the background interference will affect the accuracy of keypoint detection, which will significantly deteriorate the final performance.

The factual path is designed to simulate the phenomenon of background interference.
In this path, \textcolor[RGB]{83,129,53}{$ \textbf{FPN}_f $} completes feature aggregation and generates features with the target satellite and irrelevant background:
\begin{equation}
\mathcal{F}^{f} = {\color[RGB]{83,129,53}\textbf{FPN}_f} (\textbf{F}), \label{equ1}
\end{equation}
where $\mathcal{F}^{f}$ denotes the factual feature set $\mathcal{F}^{f} = \{ \textbf{F}^{f}_n \}, (n=1,2,\cdots,N)$ generated by different layers of \textcolor[RGB]{83,129,53}{$ \textbf{FPN}_f $}. 
This feature is regarded as the total effect (TE) in counterfactual analysis.
The total direct effect (TDE) in CA-SpaceNet is replaced with TE when analyzing the factual path separately.
$ \mathcal{F}^{f} $ will be directly sent to the network head to perform keypoint detection, resulting in $ \{ (\hat{u}_k,\hat{v}_k) \}, (k=1,2,\cdots,K) $, where $K$ denotes the number of the corners of the satellite's cubic body.
3D loss $\mathcal{L}_{3D}$ and object class loss $\mathcal{L}_{cls}$ are adopted in this paper. Please refer to \cite{SwissCube} for more details about these loss functions.

\subsection{Counterfactual Path} \label{C} 
The idea of counterfactual analysis is to imagine a non-existent path, that is, to study the effect under the \textit{What If} scenario.
In space scenes, the complex satellite-earth relationship and harsh illumination conditions will cause significant changes in the background.
These factors will negatively impact the feature extraction stage and eventually lead to suboptimal results.
Therefore, we imagine what features will be generated through "\textit{what if there is no target?}" in the counterfactual path.
A path composed of a DarkNet-53 and an \textcolor[RGB]{192,0,0}{$ \textbf{FPN}_c $} is constructed to realize this assumption. 
The path in red box of Fig. \ref{fig2} shows more details. The input of the counterfactual path is $I_m$ with only background information after erasing the target through the ground-truth mask. Due to the absence of the target, the generated feature map only contains background information:
\begin{equation}
\mathcal{F}^{c} = {\color[RGB]{192,0,0}\textbf{FPN}_c} (\textbf{F}^m), \label{equ2}
\end{equation}
where $\mathcal{F}^{c}$ denotes the counterfactual feature set $\mathcal{F}^{c} = \{ \textbf{F}^{c}_n \}, (n=1,2,\cdots,N)$.

            \begin{figure} 
              \centering
              \includegraphics[width=\linewidth]{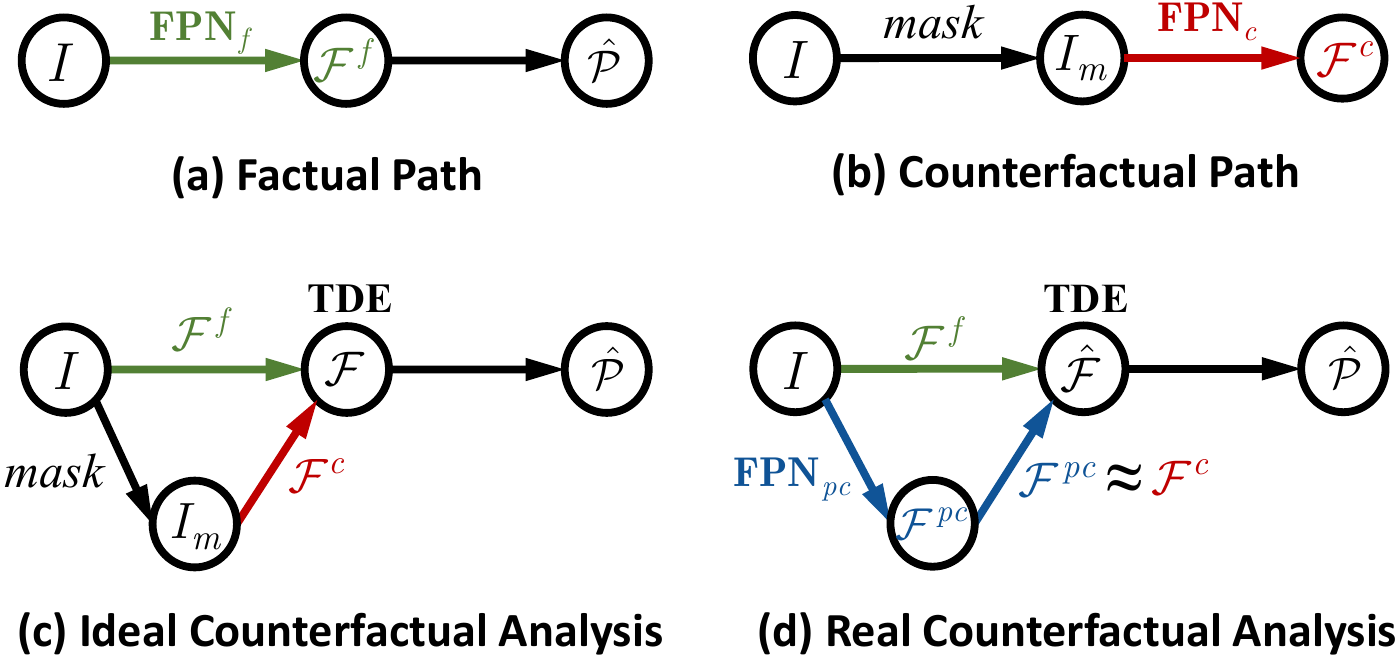}
              \caption{
              Simplified causal graphs of CA-SpaceNet in four situation. 
              These causal graphs consist of four types of nodes: image node, feature node, TDE node, and pose results node. 
              Consistent with Fig. \ref{fig2}, different colors are assigned to different elements.
              The causal graphs of the factual and counterfactual paths are shown in (a) and (b).
              The difference between the ideal (c) and the real (d) situation is caused by the unavailable masks during the inference phase.
              For clarity, $\mathcal{F}$ refers to $\mathcal{F}^{TDE}$.
              }
            \label{fig3}
            \end{figure}

Simplified causal graphs of CA-SpaceNet are shown in Fig. \ref{fig3}. 
A causal graph is a directed acyclic graph (DAG) that consists of nodes and directed edges. The nodes denote variables, and the directed edges denote cause-effects between nodes \cite{why,2016Causal}. 
Counterfactual analysis in CA-SpaceNet aims to disentangle the pure object features from the mixed features.
In Fig. \ref{fig3}, the factual path (a) and counterfactual path (b) constitute the ideal counterfactual analysis (c).
The total direct effect feature $\mathcal{F}$ in (c) is obtained by subtracting the side effect feature $\mathcal{F}^c$ from the total effect feature $\mathcal{F}^f$:
\begin{equation}
\mathcal{F} = \mathcal{F}^f - \mathcal{F}^c. \label{equ3}
\end{equation}
It should be noted that two DarkNet-53 networks in the counterfactual path and factual path share the same weights and will be frozen during training of the CA-SpaceNet. 
The reason for adopting this strategy is to only equip the FPN modules with the ability to distinguish the background and the target to avoid the backbone becoming a confounder.

\subsection{Pseudo Counterfactual Path} \label{D}
Although the unbiased feature $\mathcal{F}$ can be directly calculated by Eq. \ref{equ3} theoretically, the segmentation information of the target is not available in application.
Therefore, we elaborately design a pseudo counterfactual path to imitate the counterfactual feature $\mathcal{F}^c$, which is colored in blue in Fig. \ref{fig2}(a).
As its name implies, \textit{pseudo} means that this path is a fake path, which aims to imitate the counterfactual path.
\textcolor[RGB]{16,84,151}{$ \textbf{FPN}_{pc} $} is the main component of the pseudo counterfactual path. It takes the factual feature \textbf{F} as input and generates imitation feature:
\begin{equation}
\mathcal{F}^{pc} = {\color[RGB]{16,84,151}\textbf{FPN}_{pc}} (\textbf{F}), \label{equ4}
\end{equation}
where $\mathcal{F}^{pc}$ denotes the pseudo counterfactual feature set $\mathcal{F}^{pc} = \{ \textbf{F}^{pc}_n \}, (n=1,2,\cdots,N)$.

In order to make $\mathcal{F}^{pc}$ and $\mathcal{F}^{c}$ as similar as possible, \textit{i.e.}, $\textbf{F}^{pc}_n \approx \textbf{F}^{c}_n$, a smoothed L1 norm loss function $ sl_1(\cdot,\cdot) $ is adopted to measure the discrepancy between them:
\begin{equation}
\mathcal{L}_{sim} = \sum_{n=1}^{N} sl_1(\textbf{F}^{pc}_n, \textbf{F}^{c}_n). \label{equ5}
\end{equation}

With minimizing $\mathcal{L}_{sim}$, \textcolor[RGB]{16,84,151}{$ \textbf{FPN}_{pc} $} can learn the ability to disentangle the pure background feature from mixed feature $\textbf{F}$.
As shown in Fig. \ref{fig3}(d), the counterfactual feature $\mathcal{F}^c$ will be replaced by the pseudo counterfactual feature $\mathcal{F}^{pc}$ in real counterfactual analysis. The final approximate total direct effect feature is calculated by
\begin{equation}
\hat{\mathcal{F}} = \mathcal{F}^f - \mathcal{F}^{pc}. \label{equ6}
\end{equation}
This strategy skillfully solves the problem that the ground-truth mask is lacking during inference of the CA-SpaceNet.

            \begin{figure} 
              \centering
              \includegraphics[width=0.78 \linewidth]{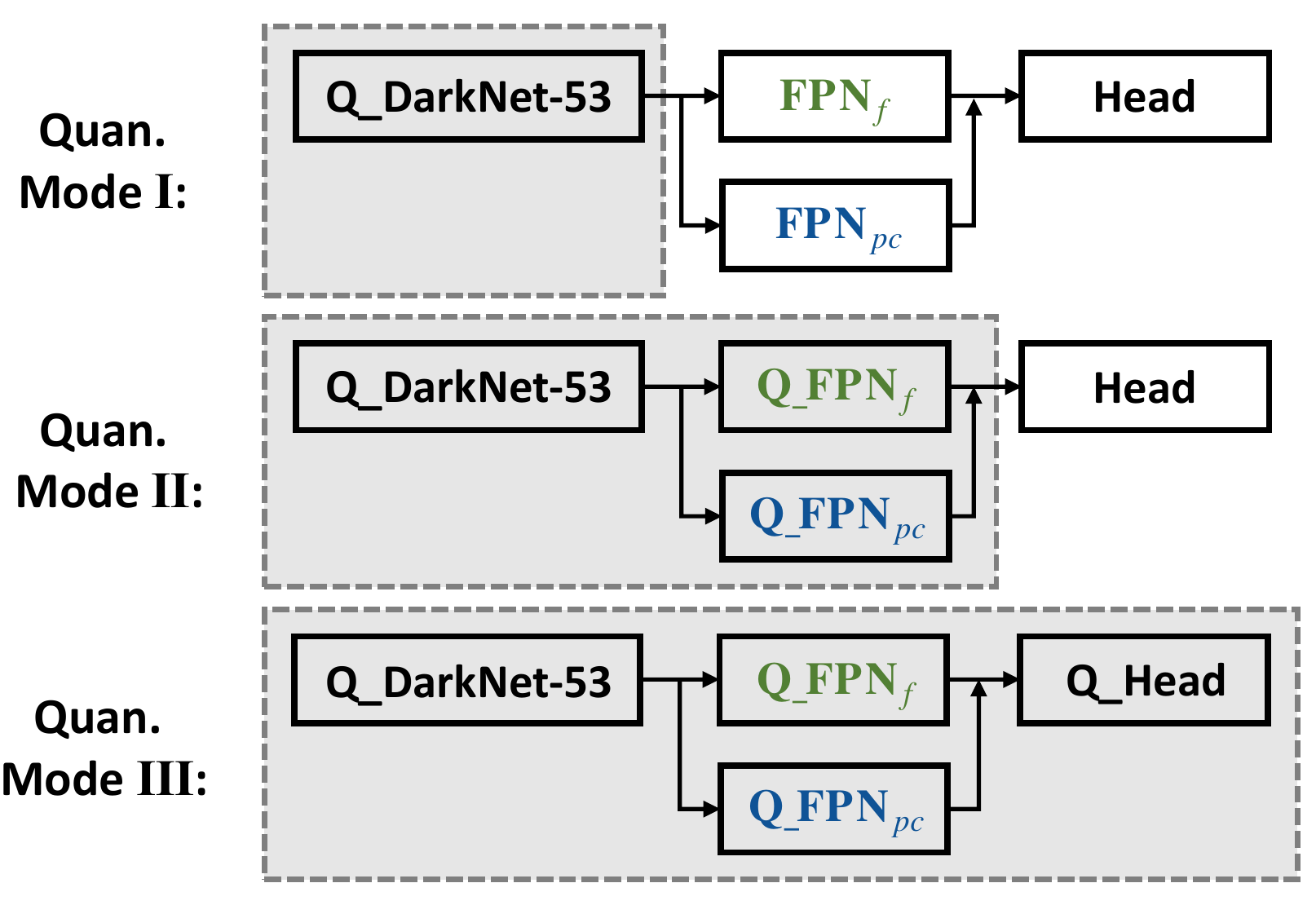}
              \caption{
              Three different quantization modes. 
              In order to explore the influence of quantizing different parts on the performance of CA-SpaceNet, three quantization modes are set up: only quantizing the backbone, quantizing the backbone and FPN, and quantizing all modules.
              Note that all counterfactual paths are removed for clarity.
              }
            \label{fig4}
            \end{figure}

\subsection{Network Quantization} \label{E}

In the space scene, special working conditions and limited computing resources put forward higher requirements for the power consumption and latency of the algorithm.
This paper adopts the classical LSQ-Net \cite{LSQ} to quantize the proposed CA-SpaceNet to a low-bit-width model and then deploys a 3-bit convolutional layer into a PIM architecture on FPGA.
As shown in Fig. \ref{fig4}, three quantization modes are proposed. The quantization range is gradually expanded to explore the impact of quantization on CA-SpaceNet finely.
This is the first work that applies network quantization methods to the 6D pose estimation task of space-borne targets.

            \begin{figure} 
              \centering
              \includegraphics[width=\linewidth]{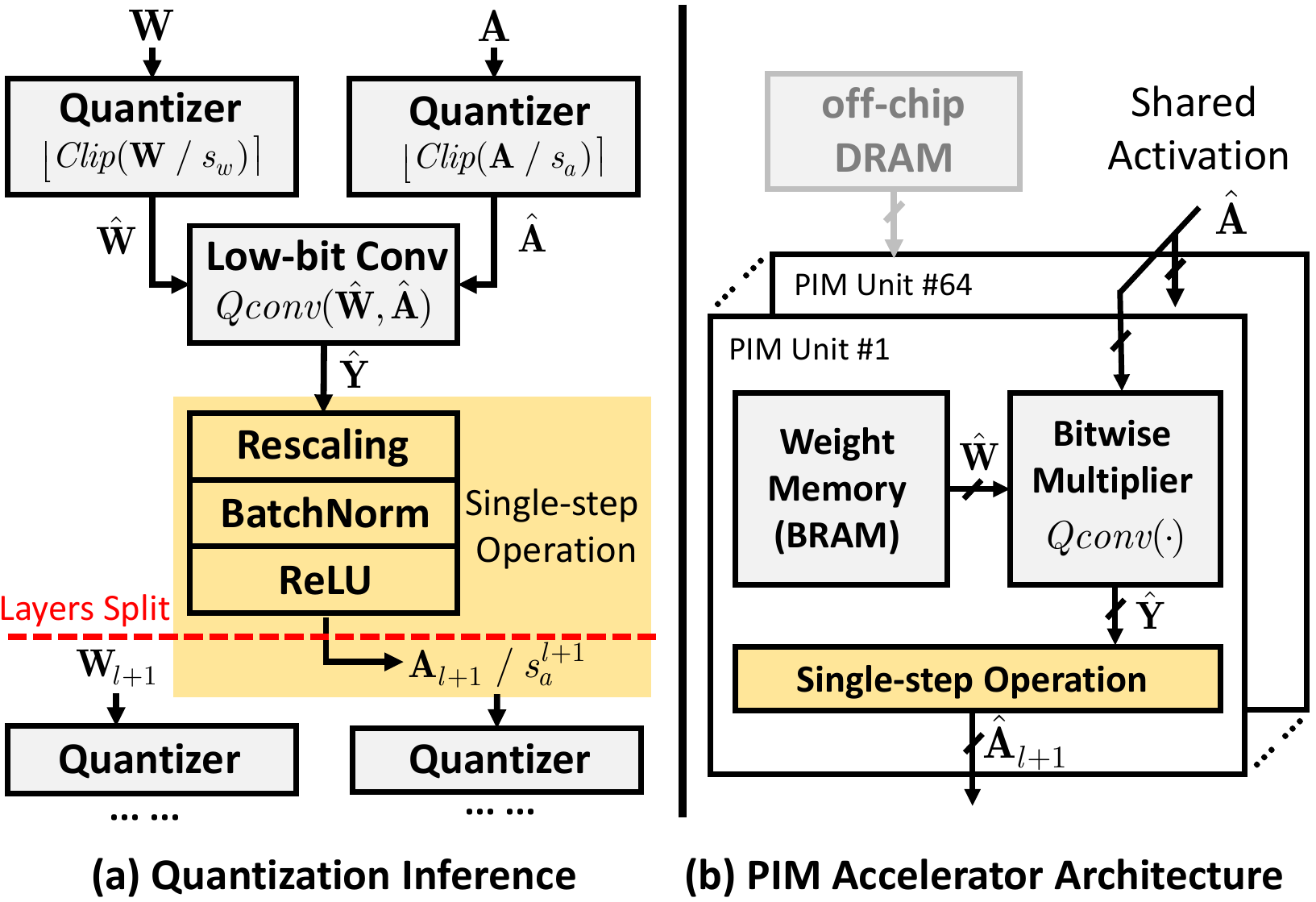}
              \caption{
              (a) Detailed quantization inference process. The red dotted line is the separation line of two Conv.-BN-Activ. layers. All computation modules in the yellow box are fused into a single-step operation.
              (b) Simplified PIM accelerator architecture adopted by this paper. Since the quantized weights have been loaded into BRAMs in 64 separate PIM units, there is no need to read data from the off-chip DRAM, which will cause low efficiency and high power consumption.
              }
            \label{fig5}
            \end{figure}

Fig. \ref{fig5}(a) demonstrates the detailed quantization and fusion process. 
Next, we will elaborate on a convolutional layer with the kernel size of $3\times 3$ and the input size of $H'\times W'$ as an example. 
Before convolution, the weights 
$\textbf{W} \in \mathbb{R}^{C_{out}\times C_{in}\times3\times3}$ 
and activations 
$\textbf{A} \in \mathbb{R}^{C_{in}\times H'\times W'} $
in FP32 are quantized into low-bit features $ \hat{\textbf{W}} = \lfloor clip(\textbf{W} / s_w) \rceil $ and $ \hat{\textbf{A}} = \lfloor clip(\textbf{A} / s_a) \rceil $
, where $ clip(\cdot) $ returns values within quantization limits, and $\lfloor \cdot \rceil$ returns every element in $\hat{\textbf{W}}$ or $\hat{\textbf{A}}$ to its nearest integer.
After the quantization convolution
\begin{equation}
  \hat{\textbf{Y}} = Qconv(\hat{\textbf{W}}, \hat{\textbf{A}}), \label{equ7}
\end{equation}
dequantization is performed to recover the activation through rescaling factors:
\begin{equation}
  \textbf{Y} = \hat{\textbf{Y}} * s_w * s_a, \label{equ8}
\end{equation}
where $\hat{\textbf{Y}}$ denotes the quantization feature and $\textbf{Y}$ denotes the recovery feature.
Then the convolutional layer and its adjacent BatchNorm layer are fused to a single operation by equivalence relation:
\begin{equation}
    \begin{aligned}
        \mathbf{Y}^{bn}_{(i,:,:,:)} &= \frac{\textbf{Y}_{(i,:,:,:)} - \bm{\mu}_i}{\bm{\sigma}_i}\bm{\gamma}_i  +\bm{\beta}_i\\
        &= \frac{\bm{\gamma}_i}{\bm{\sigma}_i} * s_w * s_a * \hat{\textbf{Y}}_{(i,:,:,:)} -\frac{\bm{\mu}_i\bm{\gamma}_i}{\bm{\sigma}_i}+\bm{\beta}_i
    \end{aligned}
    , \label{equ9}
\end{equation}
where the subscript $i$ denotes the index of the output channel $C_{out}$. $\bm{\mu}$, $\bm{\sigma}$, $\bm{\gamma}$, and $\bm{\beta}$ are four types of parameters in the BatchNorm layer.
Following this, the ReLU layer and scaling step in the next layer are also absorbed by a single operation. 


The overall weight memory occupied by the  CA-SpaceNet is greatly reduced through quantization and layer fusion, which is suitable for PIM chips with the merits of energy efficiency and avoidance of the memory wall.
In PIM architecture, the quantized network is pre-loaded into BRAM, and intermediate data accessed from/to the off-chip DRAM access is entirely eliminated during inference.
This paper implements a part of the CA-SpaceNet into the PIM accelerator proposed by Jiao \textit{et al.} \cite{Jiao}, which is demonstrated in Fig. \ref{fig5}(b). The feasibility of the deployment is confirmed on FPGA.




\subsection{Training and Inference} \label{F}
As shown in Fig. \ref{fig2}(b), the training and inference phases are separated in CA-SpaceNet. 
All three paths are activated during training. 
While learning to detect keypoints, the CA-SpaceNet is supposed to minimize the discrepancy between features generated by \textcolor[RGB]{16,84,151}{$ \textbf{FPN}_{pc} $} and \textcolor[RGB]{192,0,0}{$ \textbf{FPN}_c $}. 
Therefore, the network is trained with a weighted combination of the loss terms:
\begin{equation}
    \mathcal{L} = \lambda_{3D}\mathcal{L}_{3D} + \lambda_{c}\mathcal{L}_{cls} + \lambda_{s} \mathcal{L}_{sim},
\end{equation}
where the loss weights $\lambda_{3D}$, $\lambda_{c}$, and $\lambda_{s}$ are set to 1, 1, and 0.25, respectively. 

It should be noted that the ground-truth mask is available during training, while unavailable during inference.
The pseudo-counterfactual path is constructed to replace the function of the counterfactual path to address this issue.
Therefore, the counterfactual path (\textit{i.e.}, \textcolor[RGB]{192,0,0}{$\textbf{FPN}_c $}) will be deleted during inference. 




        \begin{table}
            \centering
            \caption{Comparison with State-of-the-arts on SwissCube.}
                \begin{tabular}{ccccc}  
                \toprule Method & Near $\uparrow$  & Medium $\uparrow$  & Far $\uparrow$  & All $\uparrow$  \\
                \midrule SegDriven \cite{hu_2019} &  41.1  &  22.9  &  7.1  &  21.8  \\
                SegDriven-Z \cite{hu_2019} &  52.6  &  45.4  &  29.4  &  43.2  \\
                DLR \cite{ICCVW_2019} &  63.8  &  47.8  &  28.9  &  46.8  \\
                WDR \cite{SwissCube} &  65.2  &  48.7  &  31.9  &  47.9  \\
                WDR* \cite{SwissCube} &  \textbf{92.37}  &  {84.16}  &  {61.27}  &  {78.78}  \\
                \midrule
                CA-SpaceNet &  {91.01}  &  \textbf{86.32}  &  \textbf{61.72}  &  \textbf{79.39}  \\
                \bottomrule
                \end{tabular}
                \label{tab1}
        \end{table}
        
        \begin{table}
            \centering
            \caption{Comparisons of the re-training WDR* model and the CA-SpaceNet.}
                \begin{tabular}{ccccc}  
                \toprule 
                Method & Near $\uparrow$  & Medium $\uparrow$  & Far $\uparrow$  & All $\uparrow$ \\
                \midrule
                WDR* \cite{SwissCube} &  \textbf{92.37}  &  84.16  &  61.27  &  78.78  \\
                \cline{2-5}
                \rule{0pt}{14pt}
                WDR* \cite{SwissCube} \textit{w.} 30-Ep.  &  \makecell[c]{89.93 \\ (-2.44)}  &  \makecell[c]{82.09 \\ (-2.07)}  &  \makecell[c]{56.50 \\ (-4.77)}  &  \makecell[c]{75.76 \\ (-3.02)}  \\
                \cline{2-5}
                \rule{0pt}{14pt}
                CA-SpaceNet & \makecell[c]{91.01 \\ (-1.36)}  &  \makecell[c]{\textbf{86.32} \\ (+2.16)}  &  \makecell[c]{\textbf{61.72} \\ (+0.45)} & \makecell[c]{\textbf{79.39} \\ (+0.61)} \\ 
                \bottomrule
                \end{tabular}
                \label{tab2}
        \end{table}

\section{Experiments}

Comprehensive experiments on SwissCube \cite{SwissCube} and SPEED \cite{SPEED} are conducted to evaluate the proposed CA-SpaceNet.
The influence of low-bit-width quantization on 6D pose estimation performance is explored. 
In the end, we deploy a layer of the quantized CA-SpaceNet to a PIM accelerator equipped on an FPGA SoC platform and evaluate the efficiency of the software and hardware co-design system.

\subsection{Datasets and Evaluation Metrics}
\noindent \textbf{SwissCube} \cite{SwissCube}. The SwissCube dataset is a high fidelity dataset for 6D object pose estimation in space scenes. Accurate 3D meshes and physically-modeled astronomical objects are included in this dataset.
It contains 500 scenes, of which each scene has 100 image sequences, resulting in 50K images in total.
Consistent with \cite{SwissCube}, 40K images are used for training, and the remaining 10K ones are used for testing.

\noindent \textbf{SPEED} \cite{SPEED}. The \textit{Spacecraft Pose Estimation Dataset} (SPEED) was firstly released on the Kelvins Satellite Pose Estimation Challenge in 2019.
It contains a large number of synthetic images and a small number of real satellite images.
The ground-truth labels of the testing set are not available because the competition is not ongoing.
Therefore, we divide the training set into two parts at random, 10K images for training and the remaining 2K ones for testing.

    \begin{figure} 
                  \centering
                  \includegraphics[width=\linewidth]{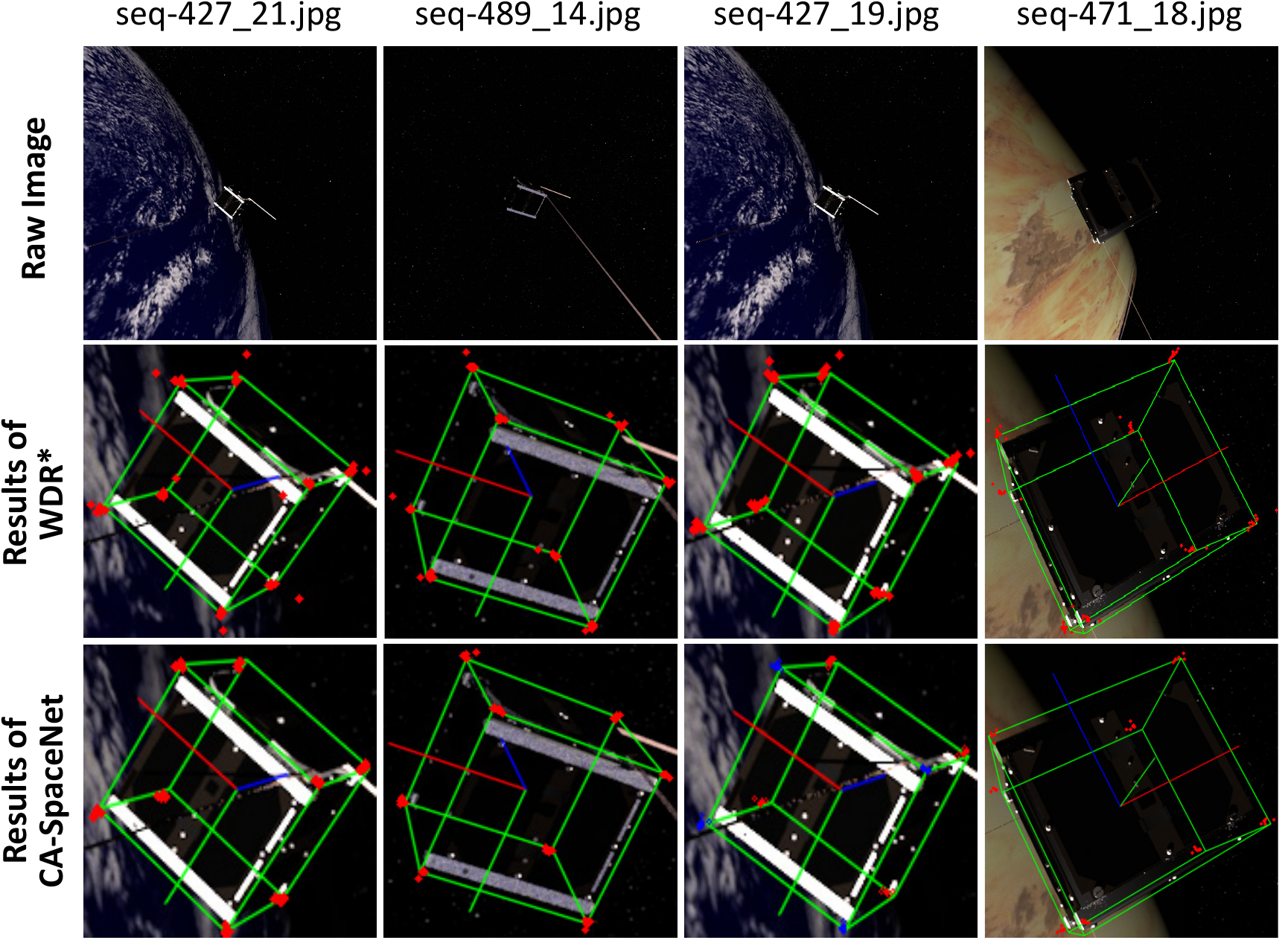}
                  \caption{
                  Visualization of the prediction keypoints generated by the WDR* and the CA-SpaceNet.
                  The ground-truth boxes (in green) and three axes are drawn for clarity.
                  All prediction points are marked in red.
                  The CA-SpaceNet can significantly reduce background interference and generate robust pose estimation results.
                  }
                \label{fig6}
    \end{figure}

\noindent \textbf{Evaluation metrics}.
Standard ADI-0.1d \cite{2020Hu,SwissCube} accuracy is adopted as the evaluation metric in SwissCube, which represents the percentage of samples whose 3D reconstruction error is less than 10\% of the object diameter.
The metric ${\textbf{e}_q} + {\textbf{e}_t}$ is adopted in SPEED, where ${\textbf{e}_q}$ is the quaternion error and ${\textbf{e}_t}$ is the normalized translation error.

\subsection{Implementation Details}
The proposed CA-SpaceNet is built on the PyTorch \cite{Pytorch} and implemented on a system with the Intel(R) Core(TM) i7-9700K CPU @ 3.60GHz. 
All methods are trained on a single Nvidia Titan GPU. 
For all CA-SpaceNet frameworks in this paper, the DarkNet-53 pretrained on ImageNet is chosen as the backbone.
Stochastic gradient descent (SGD) optimizer is adopted for network optimization with initial learning rate 1e-3, momentum 0.9, and weight decay 1e-4.
The training epoch is set to 30.
Online data augmentation strategies such as random shift, scale, and rotation are performed during training.
Different experimental settings are adopted because of different complexity.
For the SwissCube dataset, the number of minibatch is set to 8, and the resolution is set to 512×512. For the SPEED dataset, the number of minibatch is set to 4, and the resolution is set to 960×960.
An Ultra96v2 FPGA board is implemented to deploy the quantization convolutional layer of the CA-SpaceNet.

\subsection{Results on the SwissCube Dataset}
The SwissCube dataset is the largest released dataset in 6D pose estimation of space-borne targets. 
We choose this dataset as the main benchmark. 
Both quantitative and qualitative experiments are carried out on this dataset.

\noindent\textbf{Quantitative Results}.
Tab. \ref{tab1} compares the CA-SpaceNet with existing methods on the SwissCube dataset. 
Note that there are two results of WDR model \cite{SwissCube}. The results of WDR is obtained by the original paper \cite{SwissCube} while the results of WDR* is from our reproduction.
The improvement of performance lies in sufficient training epochs.
The CA-SpaceNet is obtained by introducing counterfactual analysis strategy to WDR* and performing another 30 training epochs.
Therefore, WDR* is regarded as the main competitor of the CA-SpaceNet.
In Tab. \ref{tab1}, the proposed framework achieves state-of-the-art results on ADI-0.1d (86.32 in \textit{Medium}, 61.72 in \textit{Far} and 79.39 in \textit{All}). 
Under the \textit{Medium} and \textit{Far} setting, the satellite area in the image is much smaller than the background area, which usually causes suboptimal results.
The proposed CA-SpaceNet can eliminate the interference of background through counterfactual analysis, so as to achieve better results.
This is also confirmed by subsequent qualitative experiments.
Under the \textit{Near} setting, the performance degradation is mainly caused by large masks in these scenes: the large area of the mask leads to the loss of background information, and the counterfactual path can't provide effective background features, resulting in performance degradation.

            \begin{table}
                \centering
                \caption{Comparison with State-of-the-arts on SPEED}
                    \begin{tabular}{cc}
                    \toprule Method & ${\textbf{e}_q}+{\textbf{e}_t}$ $\downarrow$ \\
                    \midrule
                    SLAB Baseline \cite{SPEED} & 0.0626 \\
                    pedro-fairspace \cite{SPEED_method_compare} & 0.0571 \\
                    WDR \cite{SwissCube} & \textbf{0.0180} \\
                    WDR* \cite{SwissCube} & 0.0400 \\
                    \midrule
                    CA-SpaceNet & 0.0385 \\
                    \bottomrule
                    \end{tabular}
                \label{tab3}
            \end{table}

In order to verify that the performance improvement is brought by the counterfactual analysis strategy rather than the additional 30 training epochs, we conducted another model named WDR* \textit{w.} 30-Ep.. 
Experimental results in Tab. \ref{tab2} show that the additional 30 training epochs lead to over-fitting, while the additional 30 training epochs with counterfactual analysis (CA-SpaceNet) achieves better performance, which confirms the superiority of the proposed framework.

\noindent\textbf{Qualitative Results}.
The prediction results of the WDR* and CA-SpaceNet are visualized and compared in Fig. \ref{fig6}.
It can be seen from the results of WDR* that the background interference makes the predicted points largely offset from the ground-truth corners. 
These imprecise corners will lead to large pose estimation errors in the PnP stage.
However, with the help of the causal inference method, the CA-SpaceNet successfully handles these complex situations.
High quality prediction points show that the counterfactual analysis strategy is able to weaken the adverse impact of background interference to the final results.

\subsection{Results on the SPEED Dataset}
Tab. \ref{tab3} compares our method with several top-performing solutions in the Kelvins Satellite Pose Estimation Challenge. 
Considering the limitation of computing resources, we didn't adopt multiple model strategies and refinement to the final 6D pose results as done in \cite{ICCVW_2019}. 
There is no official masks of the satellite in SPEED, so we utilize the \textit{cv2.convexHull} function in OpenCV to connect 11 corners to generate approximate masks.
As the original paper of WDR \cite{SwissCube} did not release the code on the SPEED dataset, we reproduce this model and obtain the results of WDR*.
Due to some unknown tricks, the results reproduced by us is slightly worse than  WDR \cite{SwissCube}, but in the same order of magnitude (0.018 for WDR and 0.040 for WDR*).
Note that this does not prevent us from evaluating the proposed framework, because the CA-SpaceNet is developed from the WDR*.
Compared with the WDR* reproduced by us, the CA-SpaceNet achieves lower error on ${\textbf{e}_q}+{\textbf{e}_t}$ (0.0385 for CA-SpaceNet and 0.04 for WDR*).
The decrease of error proves that the CA-SpaceNet is robust to complex backgrounds interference in 6D pose estimation of space-borne targets.

\begin{table}
    \centering
    \caption{Results on Three Different Quantization Modes of 8-bit and 3-bit CA-SpaceNet on SwissCube}
    \begin{tabular}{c|c|c|c|c}
		\toprule
		 \#Bits & Quan. Mode & ADI-0.1d $\uparrow$ &  OPs \& FLOPs & Perc.{(\%)}\\ 
		\midrule
		\multirow{5}{*}{8} & {\uppercase\expandafter{\romannumeral1}} & {76.21} & \makecell[c]{36.91 GOPs + \\ 33.79 GFLOPs} & {52.21}\\
		\cline{2-5}
		\rule{0pt}{14pt}
		 & {\uppercase\expandafter{\romannumeral2}} &{75.04} & \makecell[c]{44.51 GOPs + \\ 26.19 GFLOPs} &{62.96}\\
		 \cline{2-5}
		 \rule{0pt}{14pt}
		 & {\uppercase\expandafter{\romannumeral3}} &{74.65} &\makecell[c]{70.47 GOPs + \\ 0.23 GFLOPs} &{99.67}\\
		\midrule
		\multirow{5}{*}{3} & {\uppercase\expandafter{\romannumeral1}} & {75.10} & \makecell[c]{36.91 GOPs + \\ 33.79 GFLOPs} &{52.21}\\
		\cline{2-5}
		\rule{0pt}{14pt}
		 & {\uppercase\expandafter{\romannumeral2}} &{74.47} &  \makecell[c]{44.51 GOPs + \\ 26.19 GFLOPs} &{62.96}\\
		 \cline{2-5}
		 \rule{0pt}{14pt}
		 & {\uppercase\expandafter{\romannumeral3}} &{68.68} &\makecell[c]{70.47 GOPs + \\ 0.23 GFLOPs} &{99.67}\\
		\bottomrule
	\end{tabular}
	\label{tab4}
\end{table}

\subsection{Network Quantization and Deployment}
After confirming the effectiveness of the proposed framework, we quantize the CA-SpaceNet into a low-bit-width model and deploy a part of the quantized model into a real hardware accelerator.

\noindent\textbf{Quantization Results}.
We evaluate the performance of 8-bit and 3-bit CA-SpaceNet on the SwissCube dataset.
The performance and operation statistics of theses quantized models are listed in Tab. \ref{tab4}. 
The \textit{Perc.} refers to how many FLOPs in matrix multiplication are converted to low-bit-width OPs.
Details of three quantization modes are demonstrated in Fig. \ref{fig4}.
Following the common setting of quantization methods, the first layer of DarkNet-53 is kept in FP32. 
So even in mode \uppercase\expandafter{\romannumeral3}, there are still some floating-point operations that cannot be avoided.
Tab. \ref{tab4} shows that the performance of 8-bit model and 3-bit model decreases with the increase of quantization range, which is consistent with intuition.
Under the 8-bit mode \uppercase\expandafter{\romannumeral1} setting, quantifying the  DarkNet-53 can save half of the FLOPs with only reducing ADI-0.1d by 3.18 (79.39 to 76.21).
Same setting in 3-bit will reduce ADI-0.1d by 4.29 (79.39 to 75.10).
This shows that the quantization strategy can save a large mount of computation without significantly reducing the performance.
Under mode \uppercase\expandafter{\romannumeral3} setting, 8-bit and 3-bit models reduce the performance of 4.74 (79.39 to 74.65) and 10.17 (79.39 to 68.68) respectively, which shows that 3-bit quantization will have a great negative impact on the \textit{Network Head} module.
In DNNs, deeper layers represent more high-level features.
Therefore, the quantization of these modules should be firstly avoided.

Through quantization, floating-point weights in the network are transformed into low-bit-width values for storage, which greatly reduce the size of the network. 
It is more easier for the network to be deployed to devices with limited memory.
Tab. \ref{tab5} gives an occupation summary of 8-bit and 3-bit networks under mode \uppercase\expandafter{\romannumeral3} setting.
The size of the network is reduced by 75\% and 90.63\% in 8-bit and 3-bit settings, respectively.
The minimal memory occupation provides the basis for the real deployment on chips.




\noindent\textbf{Deployment on the PIM Chip}. 
Due to the limitation of hardware resources of FPGA, we only deploy a single 3-bit quantized convolutional layer of the CA-SpaceNet with the feature map size of $128\times 128\times 64$  and kernel size of $128\times 64\times 3\times 3$. 
In Tab. \ref{tab6}, the latency of PIM architecture \cite{Jiao} on FPGA is compared with the latency on ARM v8.2 CPU and Intel Core-i7 CPU. 
Note that GPU is not listed because of the low-power consumption setting.
The results show that the PIM accelerator achieves the lowest latency (5.99ms) at a clock rate of 100MHz.
Our deployment achieves 4.4x speedup compared with ARM v8.2 CPU and 1.7x speedup compared with Intel Core-i7 CPU.
Lower latency proves the high efficiency of the low-bit-width quantization and actual deployment.




\begin{table}
    \centering
    \caption{Summary of Parameter Storage Size}
        \begin{tabular}{cccc}
        \toprule Format & \#Para. & Model Size & Stor. Saving (\%) $\uparrow$ \\
        \midrule
        FP32  & 51.29 M & 205.17 MB & 0.00  \\
        8-bit & 51.29 M & 51.29 MB & 75.00  \\
        3-bit & 51.29 M & 19.23 MB & 90.63    \\
        \bottomrule
        \end{tabular}
        \label{tab5}
\end{table}

\begin{table}
    \centering
    \caption{Measured Latency on Different Hardware}
        \begin{tabular}{cc}
        \toprule Device & Latency (ms) $\downarrow$ \\
        \midrule
        ARM v8.2 64-bit CPU (Nvidia Xavier) & 26.16  \\
        Intel Core i7-8700K CPU & 10.25  \\
        \midrule
        PIM Arch. on Ultra96v2 FPGA & \textbf{5.99} \\
        \bottomrule
        \end{tabular}
        \label{tab6}
\end{table}

\section{Conclusions}
In order to address the issue that 6D pose estimation in space is vulnerable to background interference,
this paper proposes CA-SpaceNet based on counterfactual analysis to weaken the interference of background features from the mixed features.
Experimental results show that the proposed framework achieves robust performance.
Further, we quantize the CA-SpaceNet into 3-bit and 8-bit and deploy part of the quantized network to a neural network accelerator on FPGA. 
We believe that our exploration can bring new contributions to the computer vision and space technology community.
In the future, we will deploy the entire quantized network to PIM chips to better meet the demands of real space missions.







\bibliographystyle{IEEEtran}
\bibliography{IEEEexample.bib}

\begin{thebibliography}{10}
\providecommand{\url}[1]{#1}
\csname url@rmstyle\endcsname
\providecommand{\newblock}{\relax}
\providecommand{\bibinfo}[2]{#2}
\providecommand\BIBentrySTDinterwordspacing{\spaceskip=0pt\relax}
\providecommand\BIBentryALTinterwordstretchfactor{4}
\providecommand\BIBentryALTinterwordspacing{\spaceskip=\fontdimen2\font plus
\BIBentryALTinterwordstretchfactor\fontdimen3\font minus
  \fontdimen4\font\relax}
\providecommand\BIBforeignlanguage[2]{{%
\expandafter\ifx\csname l@#1\endcsname\relax
\typeout{** WARNING: IEEEtran.bst: No hyphenation pattern has been}%
\typeout{** loaded for the language `#1'. Using the pattern for}%
\typeout{** the default language instead.}%
\else
\language=\csname l@#1\endcsname
\fi
#2}}

\bibitem{iterative_1}
A.~Cropp and P.~Palmer, ``Pose estimation and relative orbit determination of a
  nearby target microsatellite using passive imagery,'' in \emph{Proc. of the
  5th Cranfield Conf. Dyn. Control Syst. Struct. Space}, 2002, p. 389–395.

\bibitem{PRISMA_1}
S.~D'Amico, M.~Benn, and J.~L. Jorgensen, ``Pose estimation of an uncooperative
  spacecraft from actual space imagery,'' \emph{International Journal of Space
  Science \& Engineering}, vol.~2, no.~2, pp. 171--188, 2014.

\bibitem{SPEED}
M.~Kisantal, S.~Sharma, T.~H. Park, D.~Izzo, and S.~D. Amico, ``Satellite pose
  estimation challenge: Dataset, competition design and results,'' \emph{IEEE
  Trans. Aero. Elec. Sys.}, vol.~56, no.~5, pp. 4083--4098, 2020.

\bibitem{SwissCube}
Y.~Hu, S.~Speierer, W.~Jakob, P.~Fua, and M.~Salzmann, ``Wide-depth-range 6d
  object pose estimation in space,'' in \emph{Proc. of the IEEE Conf. Comput.
  Vis. Pattern Recognit. (CVPR)}, 2021.

\bibitem{ICCVW_2019}
B.~Chen, J.~Cao, A.~Parra, and T.~J. Chin, ``Satellite pose estimation with
  deep landmark regression and nonlinear pose refinement,'' in \emph{Proc. of
  the IEEE Int. Conf. Comput. Vis. Workshops (ICCVW)}, 2019.

\bibitem{2016Causal}
J.~Pearl, M.~Glymour, and N.~P. Jewell, ``Causal inference in statistics: A
  primer,'' 2016.

\bibitem{why}
J.~Pearl and D.~Mackenzie, ``The book of why : the new science of cause and
  effect,'' vol. 361, no. 6405, pp. 855.2--855, 2018.

\bibitem{2017SSD}
W.~Kehl, F.~Manhardt, F.~Tombari, S.~Ilic, and N.~Navab, ``Ssd-6d: Making
  rgb-based 3d detection and 6d pose estimation great again,'' in \emph{Proc.
  of the IEEE Int. Conf. Comput. Vis. (ICCV)}, 2017.

\bibitem{2017BB8}
M.~Rad and V.~Lepetit, ``Bb8: A scalable, accurate, robust to partial occlusion
  method for predicting the 3d poses of challenging objects without using
  depth,'' in \emph{Proc. of the IEEE Int. Conf. Comput. Vis. (ICCV)}, 2017.

\bibitem{2018Real}
B.~Tekin, S.~N. Sinha, and P.~Fua, ``Real-time seamless single shot 6d object
  pose prediction,'' in \emph{Proc. of the IEEE Conf. Comput. Vis. Pattern
  Recognit. (CVPR)}, 2018.

\bibitem{PoseCNN}
Y.~Xiang, T.~Schmidt, V.~Narayanan, and D.~Fox, ``Posecnn: A convolutional
  neural network for 6d object pose estimation in cluttered scenes,'' in
  \emph{Robotics: Science and Systems Conference}, 2018.

\bibitem{PnP}
C.~P. Lu, G.~D. Hager, and E.~Mjolsness, ``Fast and globally convergent pose
  estimation from video images,'' \emph{IEEE Trans. Pattern Anal. Mach. Intell.
  (TPAMI)}, vol.~22, no.~6, pp. 610--622, 2000.

\bibitem{PoseCNN2018}
X.~Yu, T.~Schmidt, V.~Narayanan, and D.~Fox, ``Posecnn: A convolutional neural
  network for 6d object pose estimation in cluttered scenes,'' in
  \emph{Robotics: Science and Systems}, 2018.

\bibitem{ICRA_Dataset}
P.~F. Proenca and G.~Yang, ``Deep learning for spacecraft pose estimation from
  photorealistic rendering,'' in \emph{Proc. of the IEEE Int. Conf. Robot.
  Autom. (ICRA)}, 2020.

\bibitem{se3TrackNet}
B.~Wen, C.~Mitash, B.~Ren, and K.~E. Bekris, ``se(3)-tracknet: Data-driven 6d
  pose tracking by calibrating image residuals in synthetic domains,'' in
  \emph{Proc. of IEEE/RSJ Int. Conf. Intell. Robot. Syst. (IROS)}, 2020.

\bibitem{nav_sys_1}
S.~Sharma and S.~D’Amico, ``Reduced-dynamics pose estimation for
  non-cooperative spacecraft rendezvous using monocular vision,'' in
  \emph{Proc. of the AAS Guidance, Navigation and Control Conf.}, 2017.

\bibitem{nav_sys_2}
S.-G. Kim, J.~Crassidis, C.~Y., A.~Fosbury, and J.~Junkins, ``Kalman filtering
  for relative spacecraft attitude and position estimation,'' \emph{Journal of
  Guidance, Control, and Dynamics}, 2007.

\bibitem{iterative_2}
S.~Zhang and X.~Cao, ``Closed-form solution of monocular vision-based relative
  pose determination for rvd spacecrafts,'' \emph{Aircraft Engineering \&
  Aerospace Technology}, vol.~77, no.~3, pp. 192--198, 2005.

\bibitem{iterative_3}
A.~Petit, E.~Marchand, and K.~Kanani, ``Vision-based space autonomous
  rendezvous : A case study,'' in \emph{Proc. of IEEE/RSJ Int. Conf. Intell.
  Robot. Syst. (IROS)}, 2011.

\bibitem{PRISMA_2}
S.~Sumant, V.~Jacopo, and D.~Simone, ``Robust model-based monocular pose
  initialization for noncooperative spacecraft rendezvous,'' \emph{Journal of
  Spacecraft and Rockets}, vol.~55, no.~6, pp. 1414--1429, 2018.

\bibitem{PRISMA_3}
S.~D’Amico, P.~Bodin, M.~Delpech, and R.~Noteborn, ``Prisma,'' in
  \emph{Distributed Space Missions for Earth System Monitoring}, 2013.

\bibitem{Sharma_2018}
S.~Sharma, S.~D’Amico, and C.~Beierle, ``Pose estimation for non-cooperative
  spacecraft rendezvous using convolutional neural networks,'' in \emph{IEEE
  Aerospace Conference}, 2018.

\bibitem{Activism}
B.~G. King, ``A political mediation model of corporate response to social
  movement activism,'' \emph{Administrative Science Quarterly}, vol.~53, no.~3,
  pp. 395--421, 2008.

\bibitem{political}
K.~Luke, ``The statistics of causal inference: A view from political
  methodology,'' \emph{Political Analysis}, vol.~23, no.~3, pp. 313--335, 2015.

\bibitem{medical}
L.~Richiardi, R.~Bellocco, and D.~Zugna, ``Mediation analysis in epidemiology:
  methods, interpretation and bias,'' \emph{International Journal of
  Epidemiology}, vol.~42, no.~5, pp. 1511--1519, 2013.

\bibitem{Econometrica}
V.~Chernozhukov, I.~Fernández-Val, and B.~Melly, ``Inference on counterfactual
  distributions,'' \emph{Econometrica}, vol.~81, no.~6, p. 2205–2268, 2013.

\bibitem{2020Long}
K.~Tang, J.~Huang, and H.~Zhang, ``Long-tailed classification by keeping the
  good and removing the bad momentum causal effect,'' in \emph{Adv. Neural Inf.
  Process. Syst. (NeurlPS)}, 2020.

\bibitem{CA_EAA}
T.~Zhang, W.~Min, J.~Yang, T.~Liu, S.~Jiang, and Y.~Rui, ``What if we could not
  see? counterfactual analysis for egocentric action anticipation,'' in
  \emph{Proc. of the Int. Joint Conf. Artif. Intell. (IJCAI)}, 2021.

\bibitem{UnbiasedSGG}
K.~Tang, Y.~Niu, J.~Huang, J.~Shi, and H.~Zhang, ``Unbiased scene graph
  generation from biased training,'' in \emph{Proc. of the IEEE Conf. Comput.
  Vis. Pattern Recognit. (CVPR)}, 2020.

\bibitem{VQA}
K.~Tang, J.~Huang, and H.~Zhang, ``Counterfactual vqa: A cause-effect look at
  language bias,'' in \emph{Proc. of the IEEE Conf. Comput. Vis. Pattern
  Recognit. (CVPR)}, 2021.

\bibitem{pruning_1}
S.~Gao, F.~Huang, W.~Cai, and H.~Huang, ``Network pruning via performance
  maximization,'' in \emph{Proc. of the IEEE Conf. Comput. Vis. Pattern
  Recognit. (CVPR)}, 2021.

\bibitem{pruning_2}
Z.~Wang, C.~Li, and X.~Wang, ``Convolutional neural network pruning with
  structural redundancy reduction,'' in \emph{Proc. of the IEEE Conf. Comput.
  Vis. Pattern Recognit. (CVPR)}, 2021.

\bibitem{distillation1}
Y.~Zhu and Y.~Wang, ``Student customized knowledge distillation: Bridging the
  gap between student and teacher,'' in \emph{Proc. of the IEEE Int. Conf.
  Comput. Vis. (ICCV)}, 2021.

\bibitem{distillation2}
D.~Y. Park, M.-H. Cha, C.~Jeong, D.~Kim, and B.~Han, ``Learning
  student-friendly teacher networks for knowledge distillation,'' in \emph{Adv.
  Neural Inf. Process. Syst. (NeurlPS)}, 2021.

\bibitem{LSQ}
S.~K. Esser, J.~L. Mckinstry, D.~Bablani, R.~Appuswamy, and D.~S. Modha,
  ``Learned step size quantization,'' in \emph{Proc. of the Int. Conf. Learn.
  Represent. (ICLR)}, 2020.

\bibitem{quan_1}
K.~Yamamoto, ``Learnable companding quantization for accurate low-bit neural
  networks,'' in \emph{Proc. of the IEEE Conf. Comput. Vis. Pattern Recognit.
  (CVPR)}, 2021.

\bibitem{Jiao}
B.~Jiao, J.~Zhang, Y.~Xie, S.~Wang, H.~Zhu, X.~Kang, Z.~Dong, L.~Zhang, and
  C.~Chen, ``A 0.57-gops/dsp object detection pim accelerator on fpga,'' in
  \emph{26th Asia and South Pacific Design Automation Conference (ASP-DAC)},
  2021.

\bibitem{chip_review}
V.~Sze, Y.-H. Chen, T.-J. Yang, and J.~S. Emer, ``Efficient processing of deep
  neural networks: A tutorial and survey,'' \emph{Proc. of the IEEE}, vol. 105,
  no.~12, pp. 2295--2329, 2017.

\bibitem{hu_2019}
Y.~Hu, J.~Hugonot, P.~Fua, and M.~Salzmann, ``Segmentation-driven 6d object
  pose estimation,'' in \emph{Proc. of the IEEE Conf. Comput. Vis. Pattern
  Recognit. (CVPR)}, 2019.

\bibitem{2020Hu}
Y.~Hu, P.~Fua, W.~Wang, and M.~Salzmann, ``Single-stage 6d object pose
  estimation,'' in \emph{Proc. of the IEEE Conf. Comput. Vis. Pattern Recognit.
  (CVPR)}, 2020.

\bibitem{Pytorch}
A.~Paszke, S.~Gross, S.~Chintala, G.~Chanan, E.~Yang, Z.~Devito, Z.~Lin,
  A.~Desmaison, L.~Antiga, and A.~Lerer, ``Automatic differentiation in
  pytorch,'' in \emph{Adv. Neural Inf. Process. Syst. (NIPSW)}, 2017.

\bibitem{SPEED_method_compare}
P.~F. Proenca and G.~Yang, ``Deep learning for spacecraft pose estimation from
  photorealistic rendering,'' in \emph{Proc. of the IEEE Int. Conf. Robot.
  Autom. (ICRA)}, 2020.

\end{thebibliography}


\end{document}